# Potato Crop Stress Identification in Aerial Images using Deep Learning-based Object Detection

Sujata Butte[1], Aleksandar Vakanski[2*], Kasia Duellman[3], Haotian Wang[1], Amin Mirkouei[2]

[1] Department of Computer Science, University of Idaho, Idaho Falls, USA
[2] Department of Nuclear Engineering and Industrial Management, University of Idaho, Idaho Falls, USA
[3] College of Agricultural and Life Sciences, University of Idaho, Idaho Falls, USA

**Abstract:** Recent research on the application of remote sensing and deep learning-based analysis in precision agriculture demonstrated a potential for improved crop management and reduced environmental impacts of agricultural production. Despite the promising results, the practical relevance of these technologies for actual field deployment requires novel algorithms that are customized for analysis of agricultural images and robust to implementation on natural field imagery. The paper presents an approach for analyzing aerial images of a potato (*Solanum tuberosum* L.) crop using deep neural networks. The main objective is to demonstrate automated spatial recognition of a healthy versus stressed crop at a plant level. Specifically, we examine premature plant senescence resulting in drought stress on Russet Burbank potato plants. The proposed deep learning model, named Retina-UNet-Ag, is a variant of Retina-UNet (Jaeger et al., 2018) and includes connections from low-level semantic dense representation maps to the feature pyramid network. The paper also introduces a dataset of field images acquired with a Parrot Sequoia camera carried by a Solo unmanned aerial vehicle. Experimental validation demonstrated the ability for distinguishing healthy and stressed plants in field images, achieving an average Dice Score Coefficient of 0.74. A comparison to related state-of-the-art deep learning models for object detection revealed that the presented approach is effective for the task at hand. The method applied here is conducive toward the assessment and recognition of potato crop stress (early plant senescence resulting from drought stress in this case) in natural aerial field images collected under real conditions.

## INTRODUCTION

*Precision agriculture* (PA) is a concept for site-specific crop management based on observation and measurement of crop variability in a field (Pierce and Nowak, 1999; Stafford, 2000). The aim in PA is to improve crop yield and reduce the environmental impacts of agricultural production by decreased use of agrochemical substances. Despite the confirmed economic and environmental benefits of PA, professional reports and scientific surveys point to low rates of technology adoption (Pathak et al., 2019; Pierpaoli et al., 2013). Barriers towards increased adoption include the development of advanced data processing methods and platforms for automated seeding, weeding, and harvesting (Castle et al., 2016; Nahry et al., 2011). For instance, the use of cost-effective remote sensing systems in PA, such as unmanned aerial vehicles (UAVs), equipped with multispectral or hyperspectral sensors, provides an appealing means for quick observation of large fields (Maes et al., 2017; Martinelli et al., 2015; Pádua et al., 2017; Walsh et al., 2018). Intervention based on gathered sensor data and follow-up data analysis is beneficial toward reducing or avoiding economic losses (Behmann et al., 2015; Berdugo et al., 2014; Bravo et al., 2003). Similarly, the implementation of advanced machine learning algorithms for image analysis offers new insights for crop management in PA and contributes to extracting subtle patterns in acquired field imagery (Barth et al., 2018; Chen et al., 2014; Nagaraju and Chawla, 2020; Singh et al., 2018).

In the published literature, a large body of work employed image analysis via conventional machine learning approaches for crop health assessment (Barbedo, 2013; Hamuda et al., 2016). These approaches often involve an initial step of image segmentation into plant and soil pixels, typically from extracted color-index information, such as Normalized Difference Vegetation Index (NDVI), Excess Green Index (ExG), and Excess Green minus Excess Red Index (ExGR) (Bégué et al., 2018; Candiago et al., 2015; Kyratzis et al., 2017; Xue and Su, 2017). Frequently applied image analysis methods encompass support vector machines (Camargo and Smith, 2009; Rumpf et al., 2010), *k*-nearest neighbors, random forests (Lottes et al., 2017), and multivariate Gaussian classifiers (Hamuda et al., 2016). However, conventional machine learning approaches rely on manually fine-tuning a set of parameters for a given collection of images, which may lead to decreased performance on images taken under different environmental conditions (e.g., illumination, weather), growing stages, or soil types (Abade et al., 2020; Milioto et al., 2017).

A recent line of research emphasized the advantages of deep learning (DL)-based methods for image analysis in PA (Abade et al., 2020; Ghosal et al., 2018; Nagaraju and Chawla, 2020; Ramcharan et al., 2017; Taghavi Namin et al., 2018; Yamamoto et al., 2017). Contrary to the conventional methods that typically employ steps of segmentation, feature extraction, and classification, DL models comprise multi-layer architectures that integrate all processing steps into one single framework for automated feature extraction and mapping the inputs to an output function (Goodfellow et al., 2016; LeCun et al., 2015; Zhang et al., 2018). The capacity to learn rich hierarchical features at multiple levels of abstraction by DL models has been conducive to improved performance and generalization abilities, in comparison to vegetation indices and manually crafted features (Kamilaris and Prenafeta-Boldú, 2018). Several early studies for DL-based crop health assessment used convolutional neural networks (CNN) models for *classification* of healthy and diseased leaves of bananas (Amara et al., 2017), apples (Liu et al., 2018), cassava (Ramcharan et al., 2017), tomatoes (Yamamoto et al., 2017), and soybean (Ghosal et al., 2018). Similarly, Mohanty et al. (2016) applied a CNN for classification of 14 crop species and 26 diseases using a dataset of healthy and diseased plant leaves images (Hughes and Salathe, 2015).

Although DL-based classification models have proven beneficial for analysis of close-up images of leaves or crop (e.g., taken using a cellphone camera) (Mohanty et al., 2016; Sladojevic et al., 2016), recent studies shifted the emphasis toward more advanced image analysis tasks, such as image segmentation and object detection. The major advantages and practical relevance of such tasks in comparison to image classification originate from the capability to identify and localize objects (e.g., stressed areas) in agricultural field images. Image *segmentation* is the process of partitioning an image into multiple segments, by delineating the edges of the objects of interest and separating them from the background. In the context of PA, DL-based approaches were implemented for segmentation of crop and weed (Champ et al., 2020; Dyrmann et al., 2016; Gao et al., 2020; Mortensen et al., 2016; Sa et al., 2017), fruits (Zhang et al., 2020), and seeds (Toda et al., 2020). *Object detection* is a related image analysis task, where the goal is to determine whether particular objects of interest (e.g., diseased or otherwise stressed plants) are present in an image, identify the locations of all present objects of interest in the image, and determine the category (i.e., healthy vs. stressed, or specific stress type) for all found objects. In previous PA studies, a body of work employed DL for object detection with application to fruit counting (Apolo-Apolo et al., 2020; Bargoti and Underwood, 2017; Rahnemoonfar et al., 2017; Sa et al., 2016) and fruit load estimation (Koirala et al., 2019; Villacrés and Auat Cheein, 2020), disease detection (Fuentes et al., 2017; Rançon et al., 2019), and crop damage assessment (HamidiSepehr et al., 2019). The choice of applied algorithms for image segmentation versus object detection in agriculture has been primarily task-driven. Still, preparing ground truth labels for image segmentation requires demarcating all objects in images, which is substantially more time-consuming in comparison to defining the location of a set of bounding boxes in object detection. Subsequently, object detection methods have been the preferred choice in many related tasks (Abade et al., 2020; HamidiSepehr et al., 2019) where the approximate locations of the bounding boxes for the objects of interest provide a satisfactory level of accuracy.

Despite the impressive progress in recent years and the surging number of related papers and applications, there are still numerous challenges for the actual deployment of DL methods in the field. Specifically, the major obstacles involve: (a) difficulty in processing natural crop images, characterized with large crop variability across regions and growing stages, changing environmental conditions, overlapping and moving crop/plants/leaves, and cluttered background; (b) scarcity of crop datasets with corresponding labels and annotations; and (c) advanced and robust algorithms designed and optimized for processing specific crop images.

In this work, we attempt to partially address all of the above challenges. Specifically, we introduce a novel DL architecture for drought stress assessment of potato crop in natural aerial images, as well as we introduce a dataset of agricultural images. A Parrot Sequoia camera mounted on a Solo UAV flying at a low altitude of three meters above the ground was used for acquiring images of a potato field. We manually annotated the regions of healthy and stressed potato plants with rectangular bounding boxes, and posted the dataset on a dedicated website for open public access. The proposed DL architecture is a variant of the Retina-UNet (Jaeger et al., 2018) model for object detection, which employs a Feature Pyramid Network (FPN) for generating high-level feature representations for object prediction at multiple levels of abstraction. We refer to the proposed model as Retina-UNet-Ag, due to the application in agriculture. The connections between the layers in Retina-UNet-Ag are designed for detecting regions with moderate size. The experimental validation indicates ability for spatial recognition of crop stress in aerial field imagery. We conducted a performance comparison to four state-of-the-art architectures for object detection, including: Mask Region-based Convolutional NN (Mask R-CNN) (He et al., 2017), RetinaNet (Lin et al., 2020), Faster Region-based Convolutional NN (Faster R-CNN) (Ren et al., 2015), and You Only Look Once (YOLO) v3 (Redmon and Farhadi, 2017). In the comparative analysis, our proposed network produced more accurate predictions in comparison to these object detection models for the considered task.

In the published literature, a number of other studies reported research on DL-based identification of biotic crop stresses, such as water stress (An et al., 2019; King and Shellie, 2016; Ramos-Giraldo et al., 2020) and cold damage (Yang et al., 2019), as well as abiotic crop stress, such as nutrient deficiencies (Anami et al., 2020; Tran et al., 2019; Watchareeruetai et al., 2018). Several studies have also addressed the problem of DL-based crop stress localization in aerial images of natural fields taken under real conditions (Chiu et al., 2020; Zhang et al., 2019). The significance of our approach is in the design of a DL-based object detection model for potato crop stress identification in natural aerial images. Prior related DL research for potato crop assessment either focused on classification tasks using basic CNN architectures (Polder et al., 2019), tuber disease identification (Oppenheim et al., 2019), or single-leaf image classification (Pardede et al., 2018). The most similar work to ours was reported by Huang et al. (2020), which employs RetinaNet for detection of rice plants in a paddy field. However, the authors collected field images with a tractor-carried camera. Also, the comparison provided in the Results section in this paper indicates that our proposed Retina-UNet-Ag model outperformed RetinaNet (and the four other DL-models for object detection) on the potato stress detection task.

The main contributions of this paper are: (1) a novel DL model for detection of drought-stressed potato crop in natural aerial images at a plant level; and (2) an open dataset of 360 UAV images (and an augmented training dataset of 1,5000 images) of potato crop with labeled regions of healthy and stressed/diseased plants.

## MATERIALS AND METHODS

### Field and Induction of Stress

The potato field used in this study was established at the University of Idaho - Aberdeen Research and Extension Center (Bingham County, Idaho). The geographical coordinates of latitude and longitude for the Aberdeen Research Center are 42.953°N and 112.827°W, respectively. The overall area of the used potato field is 4.3 acres (1.74 hectares) and was planted to potato cv. 'Russet Burbank' on May 14, 2018; potatoes were vine-killed on September 10 and harvested on October 5. Plots were overhead irrigated with 1/8-in (3.175 mm) nozzle sprinklers spaced approximately 40 feet (12.2 m) apart, fertilized, and otherwise maintained according to recommended timing and amounts for southeast Idaho. However, plants along the first five rows along the western edge of the field experienced up to a 50% reduction in water inputs compared to the rest of the field, due to lack of additional irrigation sprinklers along the western field edge. The field was assessed for diseases and nutrient levels to ensure the stress observed was due to drought rather than other stress factors. Plants were assessed on August 13, 2018, when stressed plants due to insufficient water inputs were beginning to senesce prematurely.

### Aerial Images Collection

For collecting aerial images of the field, we employed a small UAV Solo by 3DR, shown in Figure 1. Mounted to the drone is a multispectral camera Sequoia by Parrot ("Parrot SEQUOIA+," 2017). The Sequoia camera weighs 72 grams, and it is specifically designed for use with small UAVs. It has four monochrome sensors with 1.2 MPx resolution (960×1,280 pixels) that acquire images at four wavelength bands of the electromagnetic spectrum: visible green (550 nanometers), visible red (660 nanometers), red-edge (735 nanometers), and near-infrared (790 nanometers). The sensors capture and log the reflectance within an interval of 40 nanometers around the above-listed wavelengths. The device also has an integrated RGB camera with 16 MPx resolution (3,456×4,608 pixels). Sequoia includes a down-welling irradiance sensor (also referred to as sunshine sensor) that is shown attached to the top of the drone in Figure 1. The sensor measures the level of irradiance for each of the four wavelength bands. The measurements are used for radiometric correction of reflectance images taken under different illumination conditions. The camera has an integrated inertial measurement unit (IMU) and a magnetometer, as well as a global positioning system (GPS) unit, IMU, and magnetometer built into the irradiance sensor.

The images were acquired by flying the drone at an altitude of 3 meters (9.8 feet) above the ground at a slow speed of 0.5 m/s. This altitude allows capturing high-resolution images of crop plants, and subsequently, recognition of crop stress in the field at a plant level. The ground resolution of the images is 0.08 cm for the RGB sensor, and 0.28 cm for the monochrome sensors. The footprint per image is 3.6×2.7 m for the RGB sensor, and 3.8×2.8 m for the monochrome sensors. A time-lapse mode was used for capturing the images with a time interval set to 1.21 seconds.

### Dataset

The dataset consists of labeled images containing a mix of healthy and stressed potato plants. The RGB sensor in Sequoia captured high-resolution images with 3,456×4,608 pixels size. Examples of the images are shown in Figure 2. From the collected set of high-resolution full-size RGB images, we extracted smaller image patches with 1,500×1,500 pixels size from different spatial locations by cropping and applying rotations of 45, 90, and 135 degrees. This step resulted in a set of 360 image patches, each containing several rows of healthy and stressed potato plants. The full-size images acquired with the four narrow-band multispectral sensors of Sequoia have a resolution of 960×1,280 pixels. The images taken by the monochrome sensors were first undistorted and aligned in a pre-processing step, to account for the different locations of the sensors on the device. Next, image patches with a size of 416×416 pixels were extracted for each of the four spectral bands.

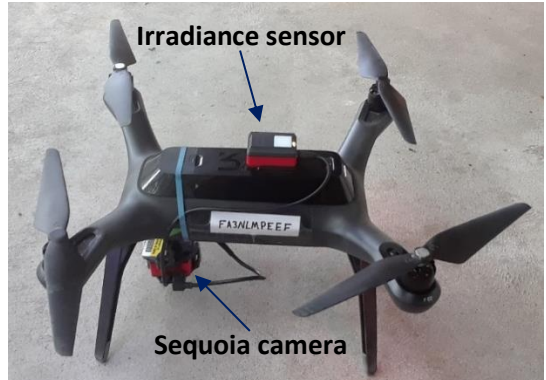

**Figure 1**. A Solo UAV and a Sequoia camera used for collecting aerial images of the field.

Besides applying translation and rotation for the extraction of image patches, we explored additional techniques to further increase the size of the training dataset, generally referred to as data augmentation. To this end, we applied the following four common methods for data augmentation: rescaling the pixels' intensity in original images (minimum percentile value = 0.2, and maximum percentile value = 99.8), adjusting the gamma value of image brightness (gamma value = 0.8, and gain value = 0.8), adjusting the sigmoid value of image contrast (cutoff value = 0.5, gain value = 10), and applying a small amount of random noise (Gaussian distribution with mean = 0 and standard distribution = 0.1). Data augmentation was applied to 300 images, resulting in an augmented training set of 1,500 images. The remaining 60 images were used for testing. Data augmentation was not applied to the test images, as well as the image patches in the test set were independent from the training set (i.e., they were extracted from different images).

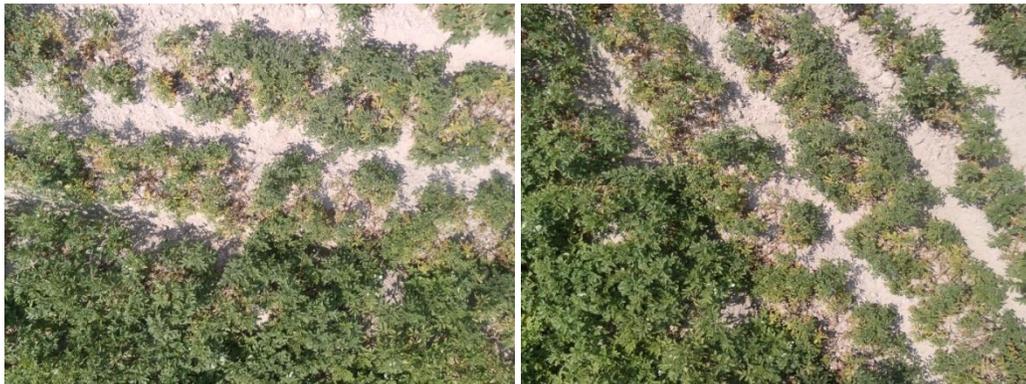

**Figure 2.** High-resolution field images with healthy and stressed plants. The stressed plants are located in the upper–half segment in the left image, and right–half segment in the right image.

Training object detection models requires labels with regions demarcating the objects of interest in images. For the potato crop dataset, the objects of interest are areas with healthy and stressed plants in images. Healthy and stressed plants were differentiated visually by color, with stressed plants exhibiting a yellower color compared to the healthy green plants. We used the open-source graphical annotation software LabelImg (Tzutalin, 2019) to manually annotate the regions containing healthy and unhealthy crop with rectangular bounding boxes. The output coordinates of the bounding boxes and the class labels were saved as XML files, and were used for generating the ground truth for model training.

The dataset is posted for open public access, and can be accessed from the following website[1].

**Deep Learning Method for Object Detection**

The proposed Retina-UNet-Ag model is a variant of the Retina-UNet architecture, which was initially proposed for object detection in medical images (Jaeger et al., 2018), and it integrates features that are characteristical for U-Net (Ronneberger et al., 2015) into the RetinaNet (Lin et al., 2020) model for object objection.

U-Net is a frequently used network for image segmentation. It employs an encoder—a contracting sequence of hidden layers (e.g., blue blocks in Figure 3) for learning semantic feature representations that gradually reduce the resolution of the feature maps, and a decoder—an expanding sequence of hidden layers (e.g., orange blocks in Figure 3) for reconstructing the feature maps to the initial size of images. A distinguishing design aspect of U-Net is the skip connections (e.g., the arrows between the blue and orange blocks in Figure 3) which enable transferring spatial information between the hidden layers in the expanding and contracting paths of the model.

RetinaNet is a well-known network for object detection, where the output consists of a set of rectangular bounding boxes for the detected objects in the images (rather than an accurate delineation of object edges, as in segmentation tasks). RetinaNet belongs to the group of one-stage object detectors, and employs a convolutional Feature Pyramid Network (FPN) (Lin et al., 2017) for generating high-level feature representations that are suitable for detecting objects at multiple levels of abstraction. While object prediction based on high-level representations is advantageous in terms of positional invariance of objects (i.e., an object can be successfully detected regardless of being at different positions across the set of images), it results in loss of important low-level semantic information, which consequently impacts the final spatial accuracy of the generated bounding boxes.

Motivated by the skip-connections in U-Net that transmit information from multiple encoding layers to multiple decoding layers (including layers with low-level feature representations, such as the layers in blocks *C4* and *C5* in Figure 3), Retina-UNet expands the FPN sub-network in RetinaNet by adding connections from the early high-resolution stages of the feature extractor from the encoding path (block *C2*) to the decoding path, as shown in Figure 3. More specifically, for a bottom-up feature extractor sub-network with stages *C2*, *C3*, *C4*, and *C5*, the top-down part of the FPN in the proposed model (orange blocks in Figure 3) is formed as:

$$P2 = Conv(P3_{upsampled} + Conv(C2)) \qquad (1)$$

$$P3 = Conv(P4_{upsampled} + Conv(C3)) \qquad (2)$$

$$P4 = Conv(P5_{upsampled} + Conv(C4)) \qquad (3)$$

$$P5 = Conv(P6_{upsampled} + Conv(C5)) \qquad (4)$$

$$P6 = Conv(C5, stride = 2) \qquad (5)$$

---

[1] https://www.webpages.uidaho.edu/vakanski/Multispectral_Images_Dataset.html

where *Conv* denotes a convolutional block layer, and the + sign signifies merging two layers by element-wise addition (e.g., the output of *P3* of size 168×168×256 is merged via element-wise addition with the output of *C2* of the same size). On the other hand, RetinaNet employs coarse feature representations from layers *P3* and above to the sub-networks for object classification and localization. Similarly, Retina-UNet architecture adds another low-level feature representation layer *P1* and another high-level layer *P7* to the FPN. For the considered crop stress identification problem, we proposed Retina-UNet-Ag to use a pyramid with top-down layers *P2* to *P6*, since there are no extremely small or large objects in comparison to the image size. Based on empirical validation, we have observed that the addition of layers *P1* and *P7* does not improve the performance of the model.

The architecture of Retina-UNet-Ag is shown in Figure 3. A ResNet-50 base model (He et al., 2016) is used for feature extraction, with input images rescaled to a size of 672×672 pixels (i.e., we selected image size dimensions that are multiples of 32). The number of the feature maps in the decoding path is 256. For detecting objects of different sizes and different width-to-height ratios, the proposed model uses anchors of sizes 16, 32, 64, 128, and 256 pixels, and ratios of 0.5, 1, and 2. The model parameters are learned by minimizing a loss function that encompasses terms for reducing the errors in class label classification and bounding box localization.

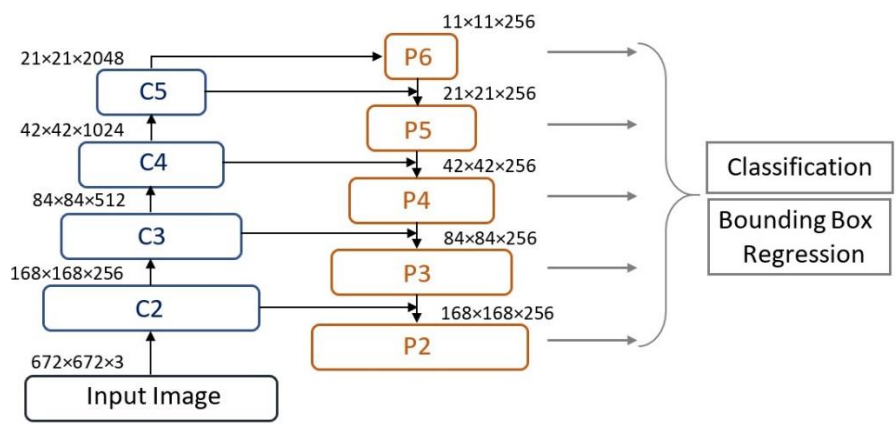

**Figure 3.** Retina-UNet-Ag architecture. *C2* to *C5* blocks output feature maps from stages 2 to 5 of the base model ResNet-50. *P2* to *P6* are the decoding block layers of the feature pyramid network, whose outputs are used for classification and bounding box regression.

**Performance Metrics**

For training and evaluation of the object detection task, we used the following metrics: Dice Score Coefficient (DSC), Intersection over Union (IoU), precision, and recall. *Dice Score Coefficient* and *Intersection over Union* measure the level of overlap between the ground truth for the objects of interest and the predictions by the model via the ratios defined respectively as $DSC = \frac{2 \cdot TP}{2 \cdot TP + FP + FN}$ and $IoU = \frac{TP}{TP + FP + FN}$. *TP* denotes the true *positives*, i.e., the set of pixels in an image that are correctly identified by the model as belonging to the positive class (either healthy or stressed plants), in accordance with the ground truth labels. *FP* denotes the *false positives*, i.e., the set of pixels in an image that are incorrectly predicted by the model to belong to the positive class. *FN* denotes the *false negatives*, i.e., the set of pixels in an image that are incorrectly predicted by the model to belong to the negative class. *Precision* calculates the percentage of the pixels in an image that are correctly predicted by the model as belonging to the positive class out of all the predicted pixels by the model, that is, Precision $= \frac{TP}{TP+FP}$. *Recall* (also known as sensitivity) expresses the percentage of the pixels that are correctly predicted as belonging to the positive class out of all relevant pixels from the positive class according to the ground truth, i.e., Recall $= \frac{TP}{TP+FN}$.

Although in most related works for object detection (Abade et al., 2020; Nagaraju and Chawla, 2020) the employed metrics—such as mean Average Precision (mAP)—are calculated based on the number of correctly predicted bounding boxes for the objects of interest, here we use metrics that are calculated based on correctly predicted pixels in crop images for healthy and stressed plants. The reason for such choice of metrics stems from the difficulty in distinguishing potato plants as single objects when the canopy is closed. In such images, a larger region with a single bounding box can contain several healthy plants, and it represents one object. An additional explanation, supported by examples, is provided in the next section.

### Training Parameters

We used the open libraries TensorFlow and Keras for the implementation of the DL model. The used hardware included both a desktop computer with a Titan XP NVIDIA GPU and the Google Colaboratory cloud computing services (offering Tesla K80 GPU). For model training, Adam optimization with a learning rate of 0.001 was used. We selected a batch size of 4 images, and a maximum number of regions of interest per image of 100. The number of training epochs was set to 15 with 500 steps per epoch, where a ResNet-50 base model was initialized with the pre-trained weights on the Common Objects in Context (COCO) dataset (Lin et al., 2014). All layers were available for parameter learning. For evaluation of the model performance, the bounding boxes with confidence scores greater than 0.7 were retained. Non-maximum suppression was applied to eliminate the bounding boxes that have an overlap greater than 30% and a lower confidence score than the neighboring bounding boxes.

## RESULTS AND DISCUSSION

Samples of images from the test set are displayed in Figure 4. The left column in Figure 4 shows four original RGB image patches collected from the field using the Sequoia camera. Manually labeled areas of healthy and drought-stressed stressed plants forming the ground truth are shown in the middle column in Figure 4. Labeled image boxes with blue edge color indicate healthy crop, and boxes with yellow edge color indicate stressed crop. In the right-hand column in Figure 4, the predicted bounding boxes by Retina-UNet-Ag are shown superimposed over the corresponding RGB images. For each bounding box, the caption in the upper left-hand corner indicates the predicted class and level of confidence. Overall, the model predicted correctly the healthy and stressed plants in almost all images.

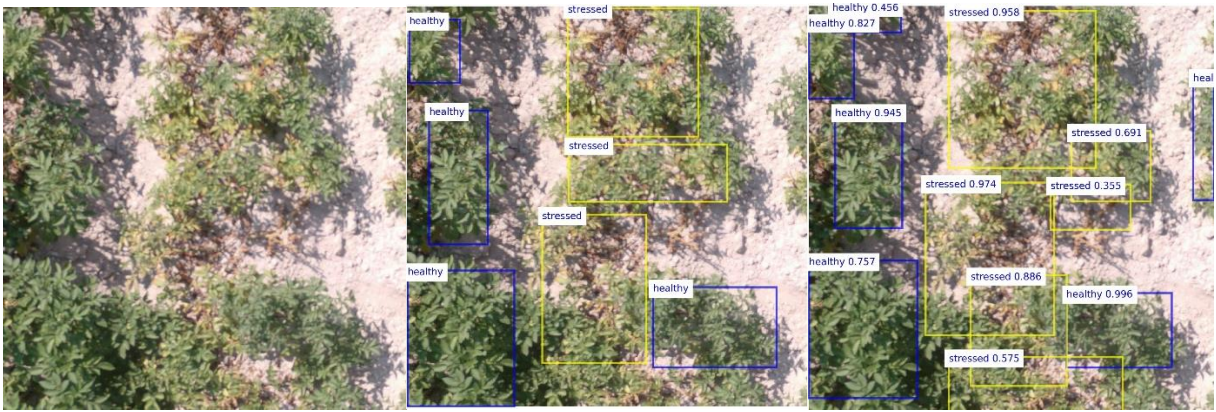

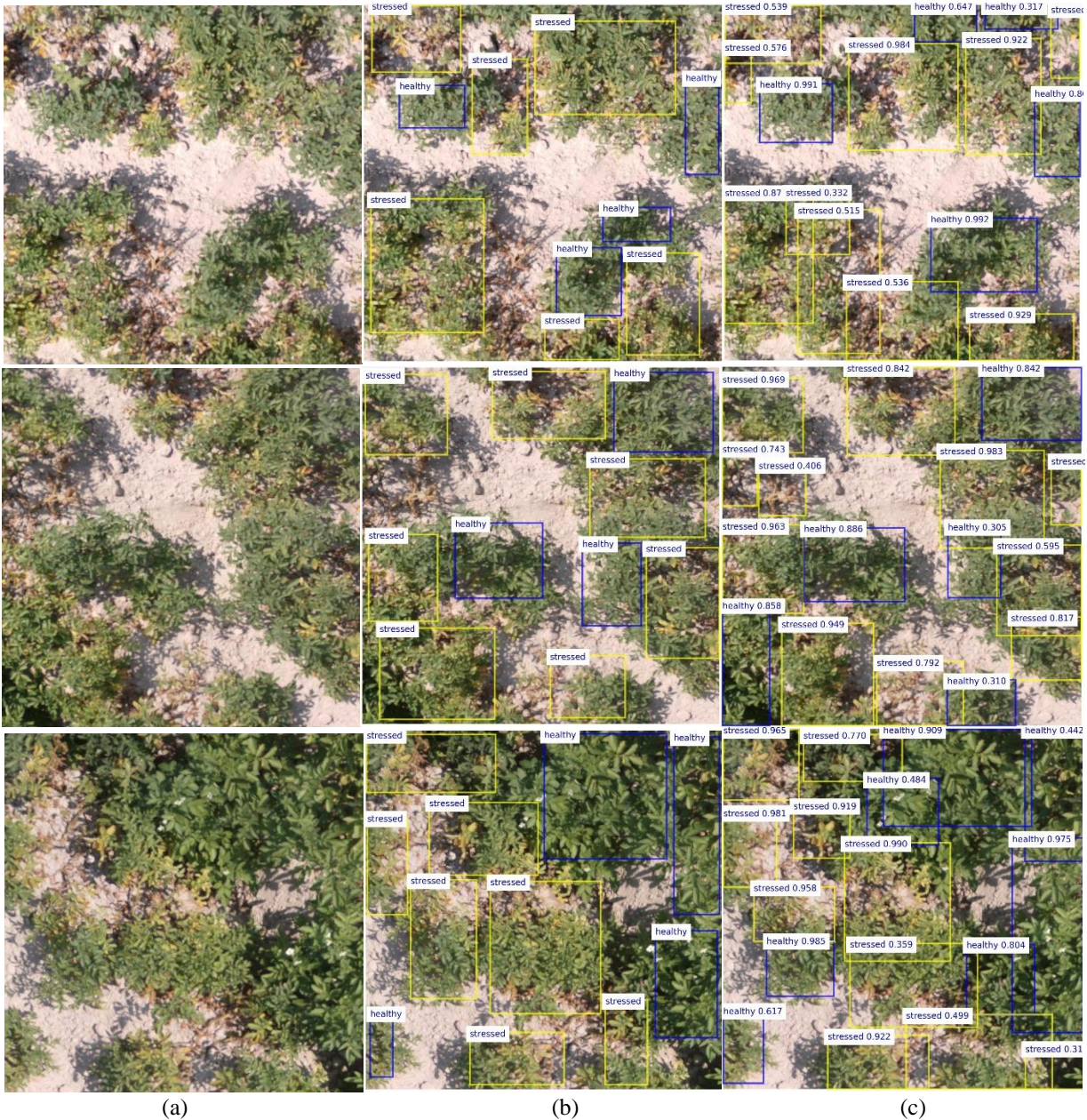

(a) (b) (c)

**Figure 4.** (a) Left column: original RGB images from the test set; (b) Middle column: ground truth; (c) Right column: Retina-UNet-Ag predictions showing the object class and level of confidence.

As noticed earlier in the text, the canopy of potato plants in the field closes toward the end of the growing season. Hence, unlike the typical objects in general object detection tasks (e.g., cars, people), the potato plants in natural field images do not have obvious boundaries to create distinct objects in the images. During the labeling of healthy and stressed crop, regions of interest demarcated with a bounding box can overlap and include several adjacent plants, as shown in the images in Figure 4. Some plants can even have a mix of healthy leaves and stressed leaves, because diseases and other stressors do not necessarily affect a whole plant uniformly. This makes labeling the collected images, and subsequently, the object detection and classification by the model, considerably challenging.

**Models Comparison**

The performance of Retina-UNet-Ag was compared to four state-of-the-art models that are commonly used for benchmarking DL architectures for object detection in general-purpose images and agricultural images (HamidiSepehr et al., 2019; Nagaraju and Chawla, 2020). The methods include Mask Region-based Convolutional NN (Mask R-CNN) with a ResNet-101 base model, RetinaNet with ResNet-50 base model, Faster Region-based Convolutional NN (Faster R-CNN) with Inception-v2 base model, and You Only Look Once (YOLO) v3 with ResNet-50 base model. All base models were pre-trained on the COCO dataset. The values of the performance metrics—DSC, IoU, Precision, and Recall—calculated on the dataset of RGB images are tabulated in Table 1. Retina-UNet-Ag achieved the highest values for both the healthy and stressed classes on almost all metrics. The average Dice Score Coefficient calculated from the values for the healthy crop ($DSC = 0.723$) and stressed crop ($DSC = 0.756$) in Table 1 amounted to 0.74. One should note that Precision and Recall are meaningful only in combination (i.e., a high value of the Precision or Recall alone does not translate into a good performance).

**Table 1.** Performance metrics values for the object detection models on the test set of RGB images.

| Model | Healthy | | | | Stressed | | | |
|---|---|---|---|---|---|---|---|---|
| | DSC | IoU | Precision | Recall | DSC | IoU | Precision | Recall |
| Retina-UNet-Ag (ours) | **0.723** | **0.574** | **0.659** | 0.832 | **0.756** | **0.604** | 0.702 | 0.841 |
| Mask R-CNN | 0.717 | 0.556 | 0.644 | 0.769 | 0.748 | 0.598 | 0.700 | 0.809 |
| RetinaNet | 0.709 | 0.537 | 0.578 | **0.899** | 0.732 | 0.583 | 0.698 | 0.795 |
| Faster R-CNN | 0.692 | 0.563 | 0.630 | 0.891 | 0.665 | 0.554 | **0.781** | 0.654 |
| Yolo v3 | 0.635 | 0.487 | 0.541 | 0.855 | 0.550 | 0.394 | 0.407 | **0.882** |

Next, the performance of Retina-UNet-Ag was evaluated on images captured with the monochrome sensors of Sequoia. The results are presented in Table 2. The evaluated images have three channels, and were created as a combination of the red (R), green (G), red-edge (RE), and near-infrared (NIR) spectral bands. The highest values for the metrics were achieved on the R-G-NIR images. Notably, the performance of the DL model for detecting healthy and stressed plants is reduced in comparison to the performance on RGB images from Table 1. As explained in the Materials section, the resolution of the four monochrome sensors of 960×1,280 pixels is significantly lower than the resolution of the RGB sensor of 3,456×4,608 pixels. A possible explanation for the decreased performance on the sets of spectral images is due to the lower resolution in connection with the altitude at which the imagery was collected. Namely, at 3 meters above the ground, the high-resolution RGB sensor captured fine details of the plants, which were conducive to distinguishing healthy from stressed leaves. More common scenarios where the use of spectral reflectance data is beneficial entail crop analysis at a field level with images taken at a height of 30 to 100 meters above the ground.

As a part of future work, we will expand the dataset with newly collected images. Additionally, our present work does not handle the type of stress or disease in the crop. Our future goal is to include classes of different stressors, and to train the model to not only identify the stressed regions, but also to identify the type of stress in the crop (e.g., distinguishing among drought, fusarium seed piece decay, and rhizoctonia canker).

Table 2. Comparison of the performance by Retina-UNet-Ag on monochrome images.

| Model | Healthy | | | | Stressed | | | |
|---|---|---|---|---|---|---|---|---|
| | DSC | IoU | Precision | Recall | DSC | IoU | Precision | Recall |
| R-G-NIR | **0.536** | 0.381 | **0.497** | 0.677 | **0.588** | **0.419** | **0.488** | 0.752 |
| R-NIR-RE | 0.529 | **0.384** | 0.466 | 0.669 | 0.582 | 0.416 | 0.480 | **0.771** |
| R-G-RE | 0.528 | 0.380 | 0.481 | **0.698** | 0.556 | 0.398 | 0.462 | 0.763 |

## CONCLUSION

The presented work introduces a method for detection and localization of drought-stressed regions in aerial images of Russet Burbank potato crop. The images were collected with a Sequoia camera airborne by a UAV flying at a low altitude of three meters above the ground. The approach employs a novel deep NN model for object detection called Retina-UNet-Ag, via introducing low-level semantically dense feature representations into the feature pyramid network of Retina-UNet. The experimental results confirm that the used approach is able to correctly identify the presence of stressed plants in almost all instances in the images. In addition, the paper introduces a dataset of aerial RGB and multispectral images with labeled regions of healthy and stressed potato crop. The dataset is posted for open access to the interested community.

## ACKNOWLEDGMENT

This work was supported through a Seed Grant awarded by the Office for Research and Economic Development (ORED) at the University of Idaho. We would like to thank Jordan Todd for his help with the collection of aerial images of the field.

## REFERENCES

Abade, A.S., Ferreira, P.A., Vidal, F. de B., 2020. Plant Diseases Recognition on Images Using Convolutional Neural Networks: A Systematic Review. Presented at the 14th International Joint Conference on Computer Vision, Imaging and Computer Graphics Theory and Applications, VISIGRAPP 2019, pp. 450-458.

Amara, J., Bouaziz, B., Algergawy, A., 2017. A Deep Learning-based Approach for Banana Leaf Diseases Classification., in: Lecture Notes in Informatics (LNI). Presented at the BTW 2017 Workshop, Bon, pp. 79–88.

An, J., Li, W., Li, M., Cui, S., Yue, H., 2019. Identification and Classification of Maize Drought Stress Using Deep Convolutional Neural Network. Symmetry 11, 256. https://doi.org/10.3390/sym11020256

Anami, B.S., Malvade, N.N., Palaiah, S., 2020. Classification of yield affecting biotic and abiotic paddy crop stresses using field images. Information Processing in Agriculture 7, 272–285. https://doi.org/10.1016/j.inpa.2019.08.005

Apolo-Apolo, O.E., Martínez-Guanter, J., Egea, G., Raja, P., Pérez-Ruiz, M., 2020. Deep learning techniques for estimation of the yield and size of citrus fruits using a UAV. European Journal of Agronomy 115, 126030. https://doi.org/10.1016/j.eja.2020.126030

Barbedo, J.G.A., 2013. Digital image processing techniques for detecting, quantifying and classifying plant diseases. Springerplus 2, 660. https://doi.org/10.1186/2193-1801-2-660

Bargoti, S., Underwood, J., 2017. Deep fruit detection in orchards, in: 2017 IEEE International Conference on Robotics and Automation (ICRA). Presented at the 2017 IEEE International Conference on Robotics and Automation (ICRA), pp. 3626–3633. https://doi.org/10.1109/ICRA.2017.7989417

Barth, R., IJsselmuiden, J., Hemming, J., Henten, E.J.V., 2018. Data synthesis methods for semantic segmentation in agriculture: A Capsicum annuum dataset. Computers and Electronics in Agriculture 144, 284–296. https://doi.org/10.1016/j.compag.2017.12.001

Bégué, A., Arvor, D., Bellón, B., Betbeder, J., Abelleyra, D. de, Ferraz, R.P.D., Lebourgeois, V., Lelong, C., Simões, M., Verón, S.R., 2018. Remote Sensing and Cropping Practices: A Review. Remote. Sens. 10, 99. https://doi.org/10.3390/rs10010099

Behmann, J., Mahlein, A.-K., Paulus, S., Kuhlmann, H., Oerke, E.-C., Plümer, L., 2015. Calibration of hyperspectral close-range pushbroom cameras for plant phenotyping. ISPRS Journal of Photogrammetry and Remote Sensing 106, 172–182. https://doi.org/10.1016/j.isprsjprs.2015.05.010

Berdugo, C.A., Zito, R., Paulus, S., Mahlein, A.-K., 2014. Fusion of sensor data for the detection and differentiation of plant diseases in cucumber. Plant Pathology 63, 1344–1356. https://doi.org/10.1111/ppa.12219

Bravo, C., Moshou, D., West, J., McCartney, A., Ramon, H., 2003. Early Disease Detection in Wheat Fields using Spectral Reflectance. Biosystems Engineering 84, 137–145. https://doi.org/10.1016/S1537-5110(02)00269-6

Camargo, A., Smith, J.S., 2009. Image pattern classification for the identification of disease causing agents in plants. Computers and Electronics in Agriculture 66, 121–125. https://doi.org/10.1016/j.compag.2009.01.003

Candiago, S., Remondino, F., Giglio, M.D., Dubbini, M., Gattelli, M., 2015. Evaluating Multispectral Images and Vegetation Indices for Precision Farming Applications from UAV Images. Remote. Sens. 7, 4026–4047. https://doi.org/10.3390/rs70404026

Castle, M.H., Lubben, B.D., Luck, J.D., 2016. Factors Influencing the Adoption of Precision Agriculture Technologies by Nebraska Producers, Digital Commons at University of Nebraska – Lincoln, pp. 1–25.

Champ, J., Mora-Fallas, A., Goëau, H., Mata-Montero, E., Bonnet, P., Joly, A., 2020. Instance segmentation for the fine detection of crop and weed plants by precision agricultural robots. Applied Plant Science 8. https://doi.org/10.1002/aps3.11373

Chen, Y., Lin, Z., Zhao, X., Wang, G., Gu, Y., 2014. Deep Learning-Based Classification of Hyperspectral Data. IEEE J. Sel. Top. Appl. Earth Obs. Remote. Sens. 7, 2094–2107. https://doi.org/10.1109/JSTARS.2014.2329330

Chiu, M.T., Xu, X., Wei, Y., Huang, Z., Schwing, A.G., Brunner, R., Khachatrian, H., Karapetyan, H., Dozier, I., Rose, G., Wilson, D., Tudor, A., Hovakimyan, N., Huang, T.S., Shi, H., 2020. Agriculture-Vision: A Large Aerial Image Database for Agricultural Pattern Analysis. CoRR abs/2001.01306.

Dyrmann, M., Mortensen, A.K., Midtiby, H.S., Jørgensen, R.N., 2016. Pixel-wise classification of weeds and crop in images by using a Fully Convolutional neural network, in: International Conference on Agricultural Engineering. Presented at the International Conference on Agricultural Engineering, Aarhus, Denmark.

Fuentes, A., Yoon, S., Kim, S.C., Park, D.S., 2017. A Robust Deep-Learning-Based Detector for Real-Time Tomato Plant Diseases and Pests Recognition. Sensors 17, 2022. https://doi.org/10.3390/s17092022

Gao, J., French, A.P., Pound, M.P., He, Y., Pridmore, T.P., Pieters, J.G., 2020. Deep convolutional neural networks for image-based Convolvulus sepium detection in sugar beet fields. Plant Methods 16, 29. https://doi.org/10.1186/s13007-020-00570-z

Ghosal, S., Blystone, D., Singh, A.K., Ganapathysubramanian, B., Singh, A., Sarkar, S., 2018. An explainable deep machine vision framework for plant stress phenotyping. Proc. Natl. Acad. Sci. U.S.A. 115, 4613–4618. https://doi.org/10.1073/pnas.1716999115

Goodfellow, I., Bengio, Y., Courville, A., 2016. Deep Learning. The MIT Press.

HamidiSepehr, A., Mirnezami, S.V., Ward, J., 2019. Comparison of object detection methods for crop damage assessment using deep learning. CoRR abs/1912.13199.

Hamuda, E., Glavin, M., Jones, E., 2016. A survey of image processing techniques for plant extraction and segmentation in the field. Computers and Electronics in Agriculture 125, 184–199. https://doi.org/10.1016/j.compag.2016.04.024

He, K., Gkioxari, G., Dollár, P., Girshick, R., 2017. Mask R-CNN, in: 2017 IEEE International Conference on Computer Vision (ICCV). Presented at the 2017 IEEE International Conference on Computer Vision (ICCV), pp. 2980–2988. https://doi.org/10.1109/ICCV.2017.322

He, K., Zhang, X., Ren, S., Sun, J., 2016. Deep Residual Learning for Image Recognition, in: 2016 IEEE Conference on Computer Vision and Pattern Recognition, CVPR 2016, Las Vegas, NV, USA, June 27-30, 2016. IEEE Computer Society, pp. 770–778. https://doi.org/10.1109/CVPR.2016.90

Huang, S., Wu, S., Sun, C., Ma, X., Jiang, Y., Qi, L., 2020. Deep localization model for intra-row crop detection in paddy field. Computers and Electronics in Agriculture, 169, 105203. https://doi.org/10.1016/j.compag.2019.105203

Hughes, D.P., Salathe, M., 2015. An open access repository of images on plant health to enable the development of mobile disease diagnostics. arXiv:1511.08060 [cs].

Jaeger, P.F., Kohl, S.A.A., Bickelhaupt, S., Isensee, F., Kuder, T.A., Schlemmer, H.-P., Maier-Hein, K.H., 2020. Retina U-Net: Embarrassingly Simple Exploitation of Segmentation Supervision for Medical Object Detection. Presented at the Machine Learning for Health NeurIPS Workshop, PMLR 116, pp. 171–183.

Kamilaris, A., Prenafeta-Boldú, F.X., 2018. Deep learning in agriculture: A survey. Computers and Electronics in Agriculture 147, 70–90. https://doi.org/10.1016/j.compag.2018.02.016

King, B.A., Shellie, K.C., 2016. Evaluation of neural network modeling to predict non-water-stressed leaf temperature in wine grape for calculation of crop water stress index. Agricultural Water Management 167, 38–52. https://doi.org/10.1016/j.agwat.2015.12.009

Koirala, A., Walsh, K.B., Wang, Z., McCarthy, C., 2019. Deep learning for real-time fruit detection and orchard fruit load estimation: benchmarking of 'MangoYOLO.' Precision Agriculture 20, 1107–1135. https://doi.org/10.1007/s11119-019-09642-0

Kyratzis, A.C., Skarlatos, D.P., Menexes, G.C., Vamvakousis, V.F., Katsiotis, A., 2017. Assessment of Vegetation Indices Derived by UAV Imagery for Durum Wheat Phenotyping under a Water Limited and Heat Stressed Mediterranean Environment. Frontiers in Plant Science 8. https://doi.org/10.3389/fpls.2017.01114

LeCun, Y., Bengio, Y., Hinton, G., 2015. Deep learning. Nature 521, 436–444. https://doi.org/10.1038/nature14539

Lin, T., Dollár, P., Girshick, R., He, K., Hariharan, B., Belongie, S., 2017. Feature Pyramid Networks for Object Detection, in: 2017 IEEE Conference on Computer Vision and Pattern Recognition (CVPR). Presented at the 2017 IEEE Conference on Computer Vision and Pattern Recognition (CVPR), pp. 936–944. https://doi.org/10.1109/CVPR.2017.106

Lin, T.-Y., Goyal, P., Girshick, R.B., He, K., Dollár, P., 2020. Focal Loss for Dense Object Detection. IEEE Transactions on Pattern Analysis and Machine Intelligence. 42, 318–327. https://doi.org/10.1109/TPAMI.2018.2858826

Lin, T.-Y., Maire, M., Belongie, S.J., Hays, J., Perona, P., Ramanan, D., Dollár, P., Zitnick, C.L., 2014. Microsoft COCO: Common Objects in Context., in: Computer Vision - ECCV 2014 - 13th European Conference, Zurich, Switzerland, September 6-12, 2014, Proceedings, Part V. pp. 740–755. https://doi.org/10.1007/978-3-319-10602-1_48

Liu, B., Zhang, Y., He, D., Li, Y., 2018. Identification of Apple Leaf Diseases Based on Deep Convolutional Neural Networks. Symmetry 10, 11. https://doi.org/10.3390/sym10010011

Lottes, P., Hörferlin, M., Sander, S., Stachniss, C., 2017. Effective Vision-based Classification for Separating Sugar Beets and Weeds for Precision Farming. Journal of Field Robotics 34, 1160–1178. https://doi.org/10.1002/rob.21675

Maes, W., Huete, A., Steppe, K., Maes, W.H., Huete, A.R., Steppe, K., 2017. Optimizing the Processing of UAV-Based Thermal Imagery. Remote Sensing 9, 476. https://doi.org/10.3390/rs9050476

Martinelli, F., Scalenghe, R., Davino, S., Panno, S., Scuderi, G., Ruisi, P., Villa, P., Stroppiana, D., Boschetti, M., Goulart, L.R., Davis, C.E., Dandekar, A.M., 2015. Advanced methods of plant disease detection. A review. Agronomy for Sustainable Development. 35, 1–25. https://doi.org/10.1007/s13593-014-0246-1

Milioto, A., Lottes, P., Stachniss, C., 2017. Real-time Semantic Segmentation of Crop and Weed for Precision Agriculture Robots Leveraging Background Knowledge in CNNs. arXiv:1709.06764 [cs].

Mohanty, S.P., Hughes, D.P., Salathé, M., 2016. Using Deep Learning for Image-Based Plant Disease Detection. Frontiers in Plant Science. 7. https://doi.org/10.3389/fpls.2016.01419

Mortensen, A.K., Dyrmann, M., Jørgensen, R.N., Gislum, R., 2016. Semantic Segmentation of Mixed Crops using Deep Convolutional Neural Network - Research - Aarhus University. Aarhus, Denmark.

Nagaraju, M., Chawla, P., 2020. Systematic review of deep learning techniques in plant disease detection. International Journal on System Assurance and Engineering Management 11, 547–560. https://doi.org/10.1007/s13198-020-00972-1

Nahry, A.H.E., Ali, R.R., Baroudy, A.A.E., 2011. An approach for precision farming under pivot irrigation system using remote sensing and GIS techniques. Agricultural Water Management 98, 517–531. https://doi.org/10.1016/j.agwat.2010.09.012

Oppenheim, D., Shani, G., Erlich, O., Tsror, L., 2019. Using Deep Learning for Image-Based Potato Tuber Disease Detection. Phytopathology 109, 1083–1087. https://doi.org/10.1094/PHYTO-08-18-0288-R

Pádua, L., Vanko, J., Hruška, J., Adão, T., Sousa, J.J., Peres, E., Morais, R., 2017. UAS, sensors, and data processing in agroforestry: a review towards practical applications. International Journal of Remote Sensing 38, 2349–2391. https://doi.org/10.1080/01431161.2017.1297548

Pardede, H.F., Suryawati, E., Sustika, R., Zilvan, V., 2018. Unsupervised Convolutional Autoencoder-Based Feature Learning for Automatic Detection of Plant Diseases, in: 2018 International Conference on Computer, Control, Informatics and Its Applications (IC3INA). Presented at the 2018 International Conference on Computer, Control, Informatics and its Applications (IC3INA), pp. 158–162. https://doi.org/10.1109/IC3INA.2018.8629518

Parrot SEQUOIA+ [WWW Document], 2017. . Parrot Store Official. URL https://www.parrot.com/business-solutions-us/parrot-professional/parrot-sequoia (accessed 9.10.19).

Pathak, H.S., Brown, P., Best, T., 2019. A systematic literature review of the factors affecting the precision agriculture adoption process. Precision Agriculture 20, 1292–1316. https://doi.org/10.1007/s11119-019-09653-x

Pierce, F.J., Nowak, P., 1999. Aspects of Precision Agriculture, in: Sparks, D.L. (Ed.), Advances in Agronomy. Academic Press, pp. 1–85. https://doi.org/10.1016/S0065-2113(08)60513-1

Pierpaoli, E., Carli, G., Pignatti, E., Canavari, M., 2013. Drivers of Precision Agriculture Technologies Adoption: A Literature Review. Procedia Technology, 6th International Conference on Information and Communication Technologies in Agriculture, Food and Environment (HAICTA 2013) 8, 61–69. https://doi.org/10.1016/j.protcy.2013.11.010

Polder, G., Blok, P.M., de Villiers, H.A.C., van der Wolf, J.M., Kamp, J., 2019. Potato Virus Y Detection in Seed Potatoes Using Deep Learning on Hyperspectral Images. Frontiers in Plant Science 10. https://doi.org/10.3389/fpls.2019.00209

Rahnemoonfar, M., Sheppard, C., Rahnemoonfar, M., Sheppard, C., 2017. Deep Count: Fruit Counting Based on Deep Simulated Learning. Sensors 17, 905. https://doi.org/10.3390/s17040905

Ramcharan, A., Baranowski, K., McCloskey, P., Ahmed, B., Legg, J., Hughes, D.P., 2017. Deep Learning for Image-Based Cassava Disease Detection. Front. Plant Sci. 8. https://doi.org/10.3389/fpls.2017.01852

Ramos-Giraldo, P., Reberg-Horton, C., Locke, A.M., Mirsky, S., Lobaton, E., 2020. Drought Stress Detection Using Low-Cost Computer Vision Systems and Machine Learning Techniques. IT Professional 22, 27–29. https://doi.org/10.1109/MITP.2020.2986103

Rançon, F., Bombrun, L., Keresztes, B., Germain, C., 2019. Comparison of SIFT Encoded and Deep Learning Features for the Classification and Detection of Esca Disease in Bordeaux Vineyards. Remote Sensing 11, 1. https://doi.org/10.3390/rs11010001

Redmon, J., Farhadi, A., 2017. YOLO9000: Better, Faster, Stronger., in: 2017 IEEE Conference on Computer Vision and Pattern Recognition, CVPR 2017, Honolulu, HI, USA, July 21-26, 2017. pp. 6517–6525. https://doi.org/10.1109/CVPR.2017.690

Ren, S., He, K., Girshick, R., Sun, J., 2015. Faster R-CNN: Towards Real-Time Object Detection with Region Proposal Networks, in: Cortes, C., Lawrence, N.D., Lee, D.D., Sugiyama, M., Garnett, R. (Eds.), Advances in Neural Information Processing Systems 28. Curran Associates, Inc., pp. 91–99.

Ronneberger, O., Fischer, P., Brox, T., 2015. U-Net: Convolutional Networks for Biomedical Image Segmentation, in: Navab, N., Hornegger, J., III, W.M.W., Frangi, A.F. (Eds.), Medical Image Computing and Computer-Assisted Intervention - MICCAI 2015 - 18th International Conference Munich, Germany, October 5 - 9, 2015, Proceedings, Part III, Lecture Notes in Computer Science. Springer, pp. 234–241. https://doi.org/10.1007/978-3-319-24574-4_28

Rumpf, T., Mahlein, A.-K., Steiner, U., Oerke, E.-C., Dehne, H.-W., Plümer, L., 2010. Early detection and classification of plant diseases with Support Vector Machines based on hyperspectral reflectance. Computers and Electronics in Agriculture 74, 91–99. https://doi.org/10.1016/j.compag.2010.06.009

Sa, I., Chen, Z., Popovic, M., Khanna, R., Liebisch, F., Nieto, J., Siegwart, R., 2017. weedNet: Dense Semantic Weed Classification Using Multispectral Images and MAV for Smart Farming. arXiv:1709.03329 [cs].

Sa, I., Ge, Z., Dayoub, F., Upcroft, B., Perez, T., McCool, C., Sa, I., Ge, Z., Dayoub, F., Upcroft, B., Perez, T., McCool, C., 2016. DeepFruits: A Fruit Detection System Using Deep Neural Networks. Sensors 16, 1222. https://doi.org/10.3390/s16081222

Singh, A.K., Ganapathysubramanian, B., Sarkar, S., Singh, A., 2018. Deep Learning for Plant Stress Phenotyping: Trends and Future Perspectives. Trends in Plant Science 23, 883–898. https://doi.org/10.1016/j.tplants.2018.07.004

Sladojevic, S., Arsenovic, M., Anderla, A., Culibrk, D., Stefanovic, D., 2016. Deep Neural Networks Based Recognition of Plant Diseases by Leaf Image Classification. Computational Intelligence and Neuroscience, 2016, 3289801:1-3289801:11. https://doi.org/10.1155/2016/3289801

Stafford, J.V., 2000. Implementing Precision Agriculture in the 21st Century. Journal of Agricultural Engineering Research 76, 267–275. https://doi.org/10.1006/jaer.2000.0577

Taghavi Namin, S., Esmaeilzadeh, M., Najafi, M., Brown, T.B., Borevitz, J.O., 2018. Deep phenotyping: deep learning for temporal phenotype/genotype classification. Plant Methods 14, 66. https://doi.org/10.1186/s13007-018-0333-4

Toda, Y., Okura, F., Ito, J., Okada, S., Kinoshita, T., Tsuji, H., Saisho, D., 2020. Training instance segmentation neural network with synthetic datasets for crop seed phenotyping. Communications Biology 3, 1–12. https://doi.org/10.1038/s42003-020-0905-5

Tran, T.-T., Choi, J.-W., Le, T.-T.H., Kim, J.-W., 2019. A Comparative Study of Deep CNN in Forecasting and Classifying the Macronutrient Deficiencies on Development of Tomato Plant. Applied Sciences 9, 1601. https://doi.org/10.3390/app9081601

Tzutalin, 2019. LabelImg. Git code: https://github.com/tzutalin/labelImg

Villacrés, J.F., Auat Cheein, F., 2020. Detection and Characterization of Cherries: A Deep Learning Usability Case Study in Chile. Agronomy 10, 835. https://doi.org/10.3390/agronomy10060835

Walsh, O.S., Shafian, S., Marshall, J.M., Jackson, C., McClintick-Chess, J.R., Blanscet, S.M., Swoboda, K., Thompson, C., Belmont, K.M., Walsh, W.L., 2018. Assessment of UAV Based Vegetation Indices for Nitrogen Concentration Estimation in Spring Wheat. Advances in Remote Sensing 07, 71. https://doi.org/10.4236/ars.2018.72006

Watchareeruetai, U., Noinongyao, P., Wattanapaiboonsuk, C., Khantiviriya, P., Duangsrisai, S., 2018. Identification of Plant Nutrient Deficiencies Using Convolutional Neural Networks, in: 2018 International Electrical Engineering Congress (IEECON). Presented at the 2018 International Electrical Engineering Congress (iEECON), pp. 1–4. https://doi.org/10.1109/IEECON.2018.8712217

Xue, J., Su, B., 2017. Significant Remote Sensing Vegetation Indices: A Review of Developments and Applications [WWW Document]. Journal of Sensors. https://doi.org/10.1155/2017/1353691

Yamamoto, K., Togami, T., Yamaguchi, N., 2017. Super-Resolution of Plant Disease Images for the Acceleration of Image-based Phenotyping and Vigor Diagnosis in Agriculture. Sensors 17, 2557. https://doi.org/10.3390/s17112557

Yang, W., Yang, C., Hao, Z., Xie, C., Li, M., 2019. Diagnosis of Plant Cold Damage Based on Hyperspectral Imaging and Convolutional Neural Network. IEEE Access 7, 118239–118248. https://doi.org/10.1109/ACCESS.2019.2936892

Zhang, C., Zou, K., Pan, Y., 2020. A Method of Apple Image Segmentation Based on Color-Texture Fusion Feature and Machine Learning. Agronomy 10, 972. https://doi.org/10.3390/agronomy10070972

Zhang, Q., Yang, L.T., Chen, Z., Li, P., 2018. A survey on deep learning for big data. Information Fusion 42, 146–157. https://doi.org/10.1016/j.inffus.2017.10.006

Zhang, X., Han, Liangxiu, Dong, Y., Shi, Y., Huang, W., Han, Lianghao, González-Moreno, P., Ma, H., Ye, H., Sobeih, T., 2019. A Deep Learning-Based Approach for Automated Yellow Rust Disease Detection from High-Resolution Hyperspectral UAV Images. Remote Sensing 11, 1554. https://doi.org/10.3390/rs11131554